\documentclass{article}

\usepackage[final]{corl_2026} 
\usepackage{graphicx}
\usepackage[utf8]{inputenc}
\usepackage{amsmath, amssymb, amsfonts} 
\usepackage{booktabs} 
\usepackage{array}
\usepackage{bm} 
\usepackage{hyperref} 
\usepackage{float}
\usepackage{wrapfig}
\usepackage{placeins}

\title{LAPT-Scale: Scaling Embodied Policies with Latent Action Pre-Training on Human Videos}

\author{
  \textbf{Zhiyuan Xiang}\textsuperscript{1,2,*} \quad
  \textbf{Peidong Liu}\textsuperscript{1,*,\ensuremath{\dagger}} \quad
  \textbf{Mingyang Li}\textsuperscript{1,3,*} \quad
  \textbf{Jiawei Li}\textsuperscript{1,\ensuremath{\dagger}} \\
  \textbf{Wenhow Li}\textsuperscript{1,4} \quad
  \textbf{Wenbo Ding}\textsuperscript{3} \quad
  \textbf{Xiang Deng}\textsuperscript{2,\ensuremath{\dagger}} \\
  {\normalfont\small \textsuperscript{1}Joy Future Academy, JD} \\
  {\normalfont\small \textsuperscript{2}School of Computer Science and Technology, Harbin Institute of Technology, Shenzhen} \\
  {\normalfont\small \textsuperscript{3}Tsinghua SIGS, Tsinghua University} \\
  {\normalfont\small \textsuperscript{4}Artificial Intelligence, The Hong Kong University of Science and Technology (Guangzhou)} \\
  {\normalfont\small
  \textsuperscript{*}Equal contribution \quad
  \textsuperscript{\ensuremath{\dagger}}Corresponding authors} \\
  {\normalfont\small
  \href{mailto:lpd19@tsinghua.org.cn}{\texttt{lpd19@tsinghua.org.cn}} \quad
  \href{mailto:li-jw15@tsinghua.org.cn}{\texttt{li-jw15@tsinghua.org.cn}} \quad
  \href{mailto:dengxiang@hit.edu.cn}{\texttt{dengxiang@hit.edu.cn}}}
}

\hypersetup{
  pdftitle={LAPT-Scale: Scaling Embodied Policies with Latent Action Pre-Training on Human Videos},
  pdfauthor={Zhiyuan Xiang, Peidong Liu, Mingyang Li, Jiawei Li, Wenhow Li, Wenbo Ding, Xiang Deng},
  hypertexnames=false
}

\begin{document}
\raggedbottom
\maketitle


\begin{abstract}
Scaling generalist robotic policies is limited by the high cost of collecting 
embodied interaction data. Large-scale egocentric human videos offer a rich 
alternative, but they lack robot action labels, and existing action annotation 
methods are often noisy under real-world visual complexity and diverse 
manipulation dynamics. We introduce \textbf{LAPT-Scale}, a scalable latent
action pre-training paradigm with two components: \textbf{LAPT-Scale-LAM}
learns continuous latent actions by explaining frame-to-frame visual transitions
in human videos and uses them as human-derived supervision, while
\textbf{LAPT-Scale-Policy} trains a VLA policy through human-video latent-action
pre-training, target-robot embodiment alignment, and task-specific
post-training. Unlike prior vector-quantized LAMs, LAPT-Scale-LAM uses a
continuous latent space to preserve fine-grained motion and interaction
dynamics. We further combine perceptual, semantic, and optical-flow alignment
signals to improve transition-level supervision beyond pixel reconstruction. Trained on
23.3K hours of egocentric video, LAPT-Scale-LAM provides implicit action
supervision for policy pre-training, yielding consistent scaling gains and
improving real-robot success from 63.3\% to 85.0\% as annotated human video
scales from 1K to 5K hours. The learned policy also generalizes to unseen
visual conditions.
\end{abstract}

\keywords{Latent action models, human-video pre-training, robot policy scaling} 


\section{Introduction}
\label{sec:intro}

Scaling data drives progress in vision and language, but embodied intelligence
faces a sharper bottleneck: interaction data requires physical robots or
simulators, making it far more expensive than text or static images. Large-scale
human videos, especially egocentric ones, offer a promising way to relieve this
bottleneck~\cite{grauman2022ego4d}. They naturally record diverse everyday
interactions at a scale beyond robot-collected trajectories, and transforming
such videos into action-like supervision could complement robot-only
datasets such as Open X-Embodiment~\cite{o2024open}.

The central challenge is that human videos lack robot action labels, and human
motion cannot be directly mapped to robot control spaces. Existing methods,
such as VITRA~\cite{li2025vitra}, often convert human videos into pseudo-actions
through explicit motion estimation, where estimation errors can lead to noisy
action annotations;
simply adding more human video does not necessarily improve policies. Latent
Action Models (LAMs) offer an alternative: they infer implicit actions between
consecutive frames end-to-end, without predefined robot action
labels~\cite{bruce2024genie,ye2024latent,bu2025univla,bi2025motus}.
Because these implicit actions are optimized to explain visual transitions, they
provide compact supervision for unlabeled videos.

Despite this potential, existing LAMs are limited by their architectural priors
and alignment signals. Many recent methods use Vector Quantized (VQ) latent
spaces~\cite{van2017neural,ye2024latent,bu2025univla}. While compact, a fixed
discrete codebook can over-compress the continuous changes that define
egocentric manipulation, weakening the supervision needed for fine-grained
policy learning. In addition, pixel-wise reconstruction signals such as MSE
are often dominated by large static regions, providing weak signals for small
but important hand-object interactions. As a result, learned latent actions may
miss transition-relevant structure needed for downstream embodied policies.

To address these limitations, we introduce \textbf{LAPT-Scale}, a scalable
latent action pre-training paradigm for embodied policy learning. LAPT-Scale
contains two components: \textbf{LAPT-Scale-LAM}, which constructs
human-derived supervision with a continuous Latent Action Model trained on
large-scale human videos, and \textbf{LAPT-Scale-Policy}, which uses this
supervision to train a VLA policy. The policy training follows three stages:
human-video latent-action pre-training, embodiment alignment with target-robot
data, and task-specific post-training in simulation and the real world. Unlike
discrete LAMs, LAPT-Scale-LAM parameterizes implicit actions in a continuous
space to preserve fine-grained transition dynamics. To make this space
perceptually grounded and temporally consistent, we train the LAM with
alignment signals beyond pixel reconstruction: perceptual alignment for local contact
fidelity~\cite{zhang2018unreasonableeffectivenessdeepfeatures}, semantic
alignment with self-supervised features for hand-object structure~\cite{oquab2023dinov2},
and optical-flow alignment for observed dynamics~\cite{teed2020raft}.

We train LAPT-Scale-LAM on 23.3K hours of diverse egocentric videos and use
it to annotate a 5.1K-hour cleaned human-video subset for policy pre-training.
With 250 hours of self-collected robot data for embodiment alignment,
LAPT-Scale reaches 90.0\%/89.3\% success on RoboTwin Easy/Hard, 64.8\% on
RoboCasa-GR1. Scaling LAM-annotated human video from 1K to 5K hours further
improves real-robot success from 63.3\% to 85.0\%, suggesting that continuous
latent action annotation is a promising route for benefiting from larger
human-video corpora. On four harder real-robot tasks, it retains 70.0--75.0\%
success under unseen position, illumination, and background shifts.

Our contributions are summarized as follows:
\begin{itemize}
\item \textbf{LAPT-Scale-LAM:} a continuous latent action model with perceptual,
self-supervised feature, and optical-flow alignment signals for learning a more
expressive latent action space from human videos.
\item \textbf{LAPT-Scale-Policy:} a three-stage framework for human-video
latent-action pre-training, target-robot embodiment alignment, and downstream
post-training.
\item \textbf{Real-robot scaling:} increasing only Stage-1 human-video
pre-training data improves downstream real-robot success from 63.3\% to 85.0\%.
\end{itemize}


\begin{figure}[t]
    \centering
    \includegraphics[width=\textwidth]{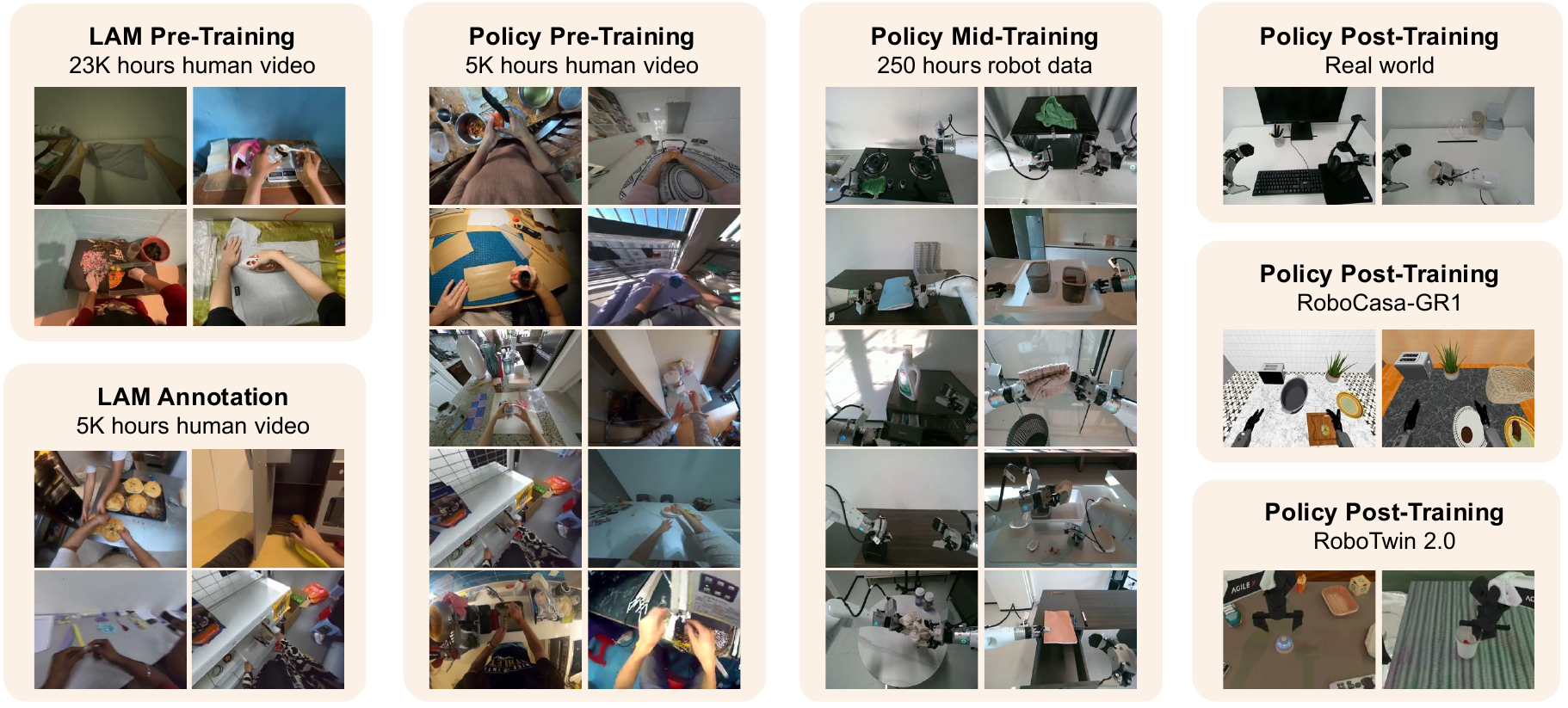}
    \caption{\textbf{Overview of LAPT-Scale.} LAPT-Scale-LAM annotates
    egocentric human videos with continuous implicit actions. LAPT-Scale-Policy
    uses these annotations for human-video pre-training, followed by
    target-robot embodiment alignment and downstream post-training.}
    \label{fig:method_overview}
\end{figure}

\section{Related Work}
\label{sec:related_work}

\paragraph{Vision-language-action policy learning.}
Vision-language-action (VLA) models unify visual perception, language
conditioning, and action generation for generalist manipulation. Recent systems
scale VLA policies with web knowledge, robot data, flow-based action modeling,
and cross-embodiment training~\cite{zitkovich2023rt,kim2024openvla,black2026pi0visionlanguageactionflowmodel,intelligence2025pi_,bjorck2025gr00t,team2025gemini,wu2026pragmatic,yang2026abot,joyai2026ra05},
while other work studies backbone, adaptation, spatial-representation, and
action-space choices~\cite{li2024robovlms,wang2024vladapter,qu2025spatialvla,zheng2025uniact}.
However, VLA training remains constrained by action-labeled robot trajectories.
Egocentric VLA methods such as VITRA and EgoScale reduce this bottleneck by
converting human videos into action-like targets before robot
alignment~\cite{li2025vitra,zheng2026egoscale}, but such targets depend on
explicit motion estimation or retargeting signals. Simulation-mediated
pipelines also convert reconstructed human demonstrations into physically
checked robot-centered trajectories~\cite{liu2026joyaisim}. LAPT-Scale instead
annotates human videos with LAM-derived implicit actions.

\paragraph{Human video data for embodied learning.}
Human video datasets provide broad coverage of physical interaction, from
object manipulation in Something-Something and EPIC-KITCHENS to daily-life
egocentric video in Ego4D~\cite{goyal2017something,damen2018scaling,grauman2022ego4d}.
Recent resources expand scale and annotation density with paired ego-exo views,
factory videos, 3D/4D annotations, dexterous demonstrations, multimodal
interaction labels, and filtered Ego4D text/action annotations~\cite{grauman2025egoexo4d,buildai2025egocentric10k,xperience_10m,hoque2025egodex,punamiya2026egoverse,wang2024egovid}.
These datasets are useful visual experience, but their human observations and
motion annotations do not directly specify robot-executable actions. LAPT-Scale learns latent actions to bridge this missing-label
gap before robot embodiment alignment.

\paragraph{Latent action models from videos.}
Latent Action Models infer action-like variables from observation transitions,
making unlabeled videos useful for behavior learning. Related work learns compact
transition variables through inverse-dynamics, world-model, or latent-action
objectives and integrates them into VLA or world-model
pipelines~\cite{bruce2024genie,ye2024latent,chen2024igor,gao2025adaworld,gao2026dreamdojo,bu2025univla,bi2025motus,chen2025villa,zhang2026clap,liu2026wla3}.
However, many LAMs rely on discrete or vector-quantized latents, which can
over-compress continuous egocentric manipulation dynamics~\cite{van2017neural,ye2024latent,bu2025univla}.
Reconstruction-only training can be dominated by static background regions.
LAOF augments reconstruction with optical-flow constraints to improve
robustness to visual distractors~\cite{bu2025laof}. LAPT-Scale-LAM
therefore uses continuous latents together with perceptual, self-supervised
semantic, and optical-flow alignment signals to preserve fine-grained implicit actions
for human-video policy pre-training~\cite{zhang2018unreasonableeffectivenessdeepfeatures,oquab2023dinov2,teed2020raft}.

\section{Method}
\label{sec:method}

\subsection{Problem Formulation \& Data}
\label{sec:problem_formulation}

\paragraph{Pipeline.}
As shown in Fig.~\ref{fig:method_overview}, LAPT-Scale contains two modules.
\textbf{LAPT-Scale-LAM} learns continuous latent actions from an egocentric
human-video corpus $\mathcal{D}_h=\{x_{1:T}^{(i)}\}$ and annotates a cleaned
subset as $\mathcal{D}_z^h=\{(\bm{c}_t^h,\bar{\bm{z}}_t^h)\}$.
\textbf{LAPT-Scale-Policy} trains a VLA policy in three stages:
human-video latent-action pre-training on $\mathcal{D}_z^h$, target-robot
embodiment alignment on $\mathcal{D}_r^{\mathrm{align}}$, and downstream
post-training on $\mathcal{D}_r^{\mathrm{task}}$.

\begin{figure}[t]
    \centering
    \includegraphics[width=\linewidth]{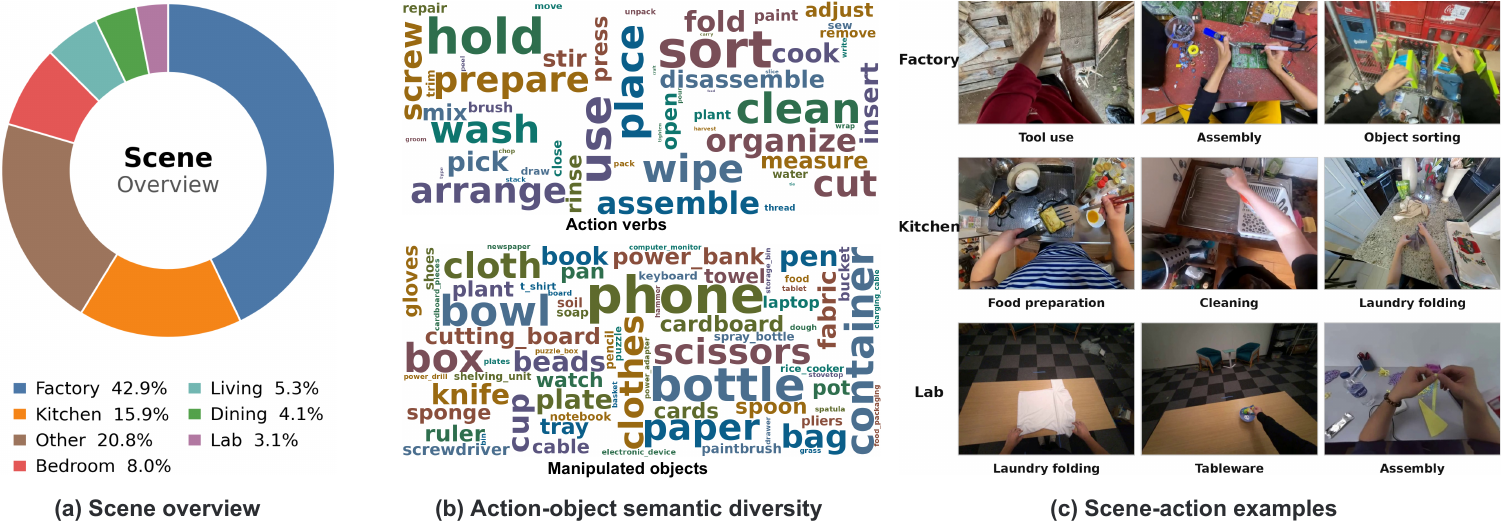}
    \caption{\textbf{Human-video data overview.} The corpus spans everyday
    egocentric manipulation across varied environments and interactions,
    providing broad coverage for latent-action learning.}
    \label{fig:data_overview}
\end{figure}

\paragraph{Human-video data and filtering.}
Our LAM is trained on 23.3K hours of diverse egocentric videos, while policy
pre-training uses a 5.1K-hour cleaned subset with more reliable manipulation
content. Fig.~\ref{fig:data_overview} summarizes the corpus diversity. Following
EgoVid-5M~\cite{wang2024egovid}, we remove corrupted or blurry clips and
discard segments dominated by camera motion or weak hand-object interaction.
This lets the LAM learn from broad visual diversity while policy scaling uses
cleaner action-like supervision.

\subsection{LAPT-Scale-LAM: Constructing Human Latent-Action Supervision}
\label{sec:lam_annotation}

\begin{wrapfigure}[20]{r}{0.48\linewidth}
    \vspace{-1.0em}
    \centering
    \includegraphics[width=\linewidth]{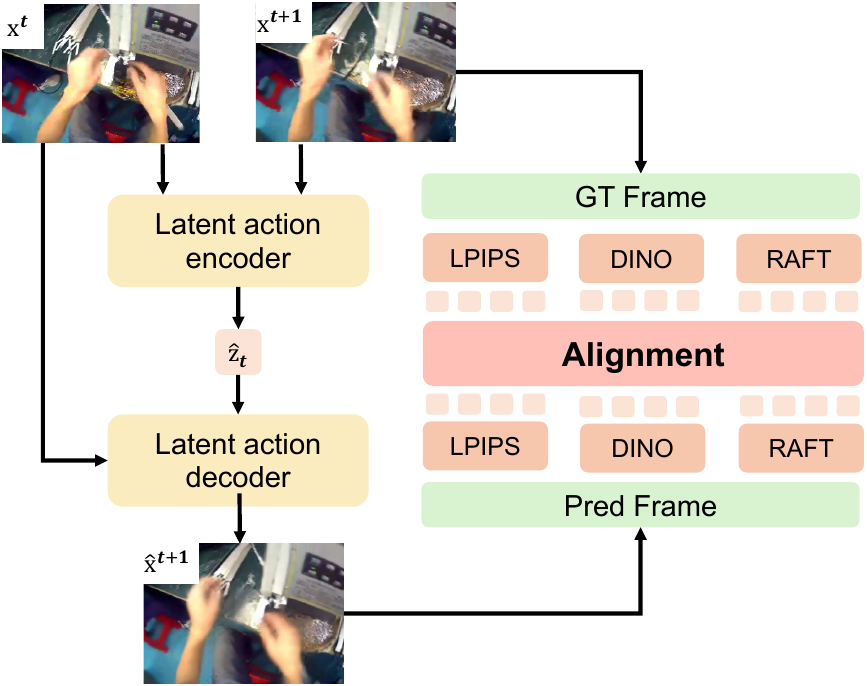}
    \caption{\textbf{LAPT-Scale-LAM training alignments.} The latent action
    explains the transition from the current frame to the next frame.
    Multi-level alignments encourage it to capture hand-object motion rather
    than static appearance.}
    \label{fig:lam_method}
\end{wrapfigure}

LAPT-Scale-LAM explains visual transitions in human videos. Given adjacent
frames $(x_t,x_{t+1})$ from a clip $x_{1:T}$, the encoder
$E_{\phi}: (x_t,x_{t+1}) \mapsto \bm{z}_t$ produces a transition-level latent
action $\bm{z}_t \in \mathbb{R}^{d}$ that captures hand-object dynamics from
$x_t$ to $x_{t+1}$. Applying the same transition encoder over the clip yields
$\bm{z}_{1:T-1}$. We implement the LAM as a spatio-temporal
VAE~\cite{kingma2013auto}: for each transition, a learnable action query
$\bm{q}_0$ is prepended to the two-frame patch embeddings and processed by a
Transformer encoder~\cite{vaswani2017attention}. The query output
parameterizes a diagonal Gaussian posterior
$q_{\phi}(\bm{z}_t \mid x_t,x_{t+1})$, from which the continuous latent action
is sampled. A low-dimensional bottleneck (e.g., $d=32$), together with a
decoder that separately receives current visual context, encourages $\bm{z}_t$
to capture frame-to-frame change rather than static appearance.

To reconstruct the future frame, we inject the latent action into current-frame
patch embeddings, $\bm{d}_{t,n}=W_p\bm{p}_{t,n}+W_z\bm{z}_t$, and use a spatial
Transformer decoder $D_{\theta}$ to predict $\hat{x}_{t+1}$. The LAM is aligned
with visual fidelity, semantic structure, and observed motion signals, as
illustrated in Fig.~\ref{fig:lam_method}. The latent action encoder
infers continuous transition variables, and the decoder is trained with
reconstruction, perceptual, semantic, and optical-flow alignment terms. These
signals are complementary: pixel and perceptual alignment keep the predicted
frame visually faithful, semantic features preserve object-centric structure
under appearance changes, and optical-flow alignment encourages the predicted transition
to match the observed motion field. We therefore supervise the decoder at
multiple levels rather than relying on reconstruction alone. Concretely, the
training alignment contains the following terms.

\begin{itemize}
    \item \textbf{Perceptual alignment.} Pixel reconstruction and
    LPIPS~\cite{zhang2018unreasonableeffectivenessdeepfeatures} preserve local structure:
    \begin{equation}
        \mathcal{L}_{\mathrm{perc}}
        = \lambda_{\mathrm{mse}}\|\hat{x}_{t+1}-x_{t+1}\|_2^2
        + \lambda_{\mathrm{lpips}}\mathrm{LPIPS}(\hat{x}_{t+1},x_{t+1}).
    \end{equation}

    \item \textbf{Semantic alignment.} Frozen DINOv2~\cite{oquab2023dinov2}
    features preserve hand-object structure:
    \begin{equation}
        \mathcal{L}_{\mathrm{dino}}
        = \frac{1}{2}\|S(P(\hat{x}_{t+1}))-S(P(x_{t+1}))\|_2^2
        + \frac{1}{2}\|c(\hat{x}_{t+1})-c(x_{t+1})\|_2^2,
    \end{equation}
    where $P$ and $c$ denote patch and CLS features, and $S(\cdot)$ denotes
    spatial pooling over patch features.

    \item \textbf{Motion alignment.} A frozen RAFT~\cite{teed2020raft}
    estimator matches reconstructed and observed optical-flow fields:
    \begin{equation}
        \mathcal{L}_{\mathrm{flow}}
        = \left\| R(x_t,\hat{x}_{t+1})
        - \mathrm{sg}[R(x_t,x_{t+1})] \right\|_1,
    \end{equation}
    where $\mathrm{sg}[\cdot]$ stops gradients through the target flow.
\end{itemize}

The full LAM training objective combines these alignment terms:
\begin{equation}
    \mathcal{L}_{\mathrm{LAM}}
    = \mathcal{L}_{\mathrm{perc}}
    + \lambda_{\mathrm{dino}}\mathcal{L}_{\mathrm{dino}}
    + \lambda_{\mathrm{flow}}\mathcal{L}_{\mathrm{flow}}
    + \beta\mathcal{L}_{\mathrm{KL}},
\end{equation}
where $\mathcal{L}_{\mathrm{KL}}$ keeps the continuous latent space close to the
prior. The frozen encoder then generates the latent-action labels used in Stage
1 of VLA training.

\subsection{LAPT-Scale-Policy: Three-Stage VLA Policy Training}
\label{sec:downstream_policy}

LAPT-Scale-Policy contains a VLM backbone $F_{\omega}$ that encodes the policy
context $\bm{c}$ and uses the JoyAI-RA policy architecture~\cite{zhang2026joyaira01foundationmodel},
with a Perceiver-style action expert $A_{\eta}$ trained by Conditional Flow
Matching~\cite{lipman2022flow}. We keep the same action expert across three
stages: Stage 1 predicts chunks of LAM-produced latent actions
$\bm{y}=\bar{\bm{z}}_{t:t+H-1}^h$ on $\mathcal{D}_z^h$, where each
$\bar{\bm{z}}_{\tau}^h$ is the posterior mean for one human-video transition
and $h$ indicates the human-video source; Stage 2 switches the target to
robot action chunks $\bm{y}=\bm{a}_{t:t+H-1}^r$ on
$\mathcal{D}_r^{\mathrm{align}}$ for embodiment alignment; Stage 3 continues on
$\mathcal{D}_r^{\mathrm{task}}$ for downstream post-training. No separate
human-video head, adapter, or reinitialization is introduced.

For each target $\bm{y}$ and context $\bm{c}$, we sample noise
$\bm{\xi}\sim\mathcal{N}(\bm{0},\mathbf{I})$ and flow time $\tau\in[0,1]$ to construct
$\bm{y}^{\tau}=\tau\bm{y}+(1-\tau)\bm{\xi}$. The flow-matching objective is
\begin{equation}
    \mathcal{L}_{\mathrm{FM}}(\bm{c},\bm{y})
    =
    \left\|
    A_{\eta}(\bm{c},\bm{y}^{\tau},\tau)-(\bm{y}-\bm{\xi})
    \right\|_2^2.
\end{equation}

Targets are padded to the action expert's maximum dimension and masked in the
loss when needed. At inference time, the policy predicts executable robot
actions; LAM-encoded actions are used only as scalable Stage-1 supervision.

\section{Experiments}
\label{sec:experiments}

We evaluate LAPT-Scale along three axes: downstream VLA performance in
simulation and the real world, scaling with LAM-annotated human-video data, and
standalone LAM quality. The downstream benchmarks test executable robot
behavior after embodiment alignment, while the LAM diagnostics test whether the
annotation model preserves transition-relevant visual and semantic information.

\subsection{Experimental Setup}
\label{sec:exp_setup}

LAPT-Scale-LAM is trained on a 23.3K-hour egocentric human-video corpus. The
main policy run uses a 5.1K-hour cleaned subset annotated by the frozen LAM,
and the scaling study uses nested 1K-, 3K-, and 5K-hour subsets with all robot
data held fixed.

For downstream evaluation, we test on RoboTwin
2.0~\cite{chen2025robotwin}, RoboCasa-GR1
Tabletop~\cite{nasiriany2024robocasa,bjorck2025gr00t}, and seven real-robot
tasks on an AgiBot G1 robot after 250 hours of target-robot embodiment
alignment. We report task success rate for policy evaluation and use
human-video reconstruction metrics plus LARY linear-probe accuracy for LAM
diagnostics. Data quantities and evaluation details are in
Appendix~\ref{sec:appendix_eval_protocol}.

\subsection{Main Results}
\label{sec:main_results}

\setlength{\intextsep}{6pt}

We compare the full LAPT-Scale pipeline against strong VLA baselines after
Stage-1 pre-training on the 5.1K-hour LAM-annotated subset, target-robot
embodiment alignment, and downstream post-training. Baseline sources and
evaluation details are summarized in the appendix.

\begin{table}[H]
\centering
\caption{Main downstream policy results on RoboTwin 2.0. Results are average
success rates in the Easy and Hard settings.}
\label{tab:main_robotwin}
\footnotesize
\setlength{\tabcolsep}{2pt}
\begin{tabular*}{\linewidth}{@{\extracolsep{\fill}}lcccccccccccc@{}}
\toprule
\textbf{Method}
& \multicolumn{2}{c}{$\pi_0$~\cite{black2026pi0visionlanguageactionflowmodel}}
& \multicolumn{2}{c}{$\pi_{0.5}$~\cite{intelligence2025pi_}}
& \multicolumn{2}{c}{Motus~\cite{bi2025motus}}
& \multicolumn{2}{c}{LingBot-VLA~\cite{wu2026pragmatic}}
& \multicolumn{2}{c}{JoyAI-RA~\cite{zhang2026joyaira01foundationmodel}}
& \multicolumn{2}{c}{\textbf{LAPT-Scale}} \\
& Easy & Hard & Easy & Hard & Easy & Hard & Easy & Hard & Easy & Hard & Easy & Hard \\
\midrule
\textbf{Success Rate (\%)} & 65.9 & 58.4 & 82.7 & 76.8 & 88.7 & 87.0 & 88.6 & 86.7 & 88.6 & 87.2 & \textbf{90.0} & \textbf{89.3} \\
\bottomrule
\end{tabular*}
\end{table}

On RoboTwin 2.0, LAPT-Scale reaches 90.0\% in the Easy setting and 89.3\% in
the Hard setting, slightly outperforming the strongest listed baselines. This
indicates that the benefit is not limited to fixed initial configurations,
since RoboTwin Hard varies object poses and scene layouts.

\begin{table}[H]
\centering
\caption{Main downstream policy results on RoboCasa-GR1. Results are average
success rates.}
\label{tab:main_robocasa}
\footnotesize
\setlength{\tabcolsep}{2pt}
\begin{tabular*}{\linewidth}{@{\extracolsep{\fill}}lccccc@{}}
\toprule
\textbf{Method} & GR00T-N1.6~\cite{bjorck2025gr00t} & Qwen3PI~\cite{community2026starvla} & ABot-M0~\cite{yang2026abot} & JoyAI-RA~\cite{zhang2026joyaira01foundationmodel} & \textbf{LAPT-Scale} \\
\midrule
\textbf{Success Rate (\%)} & 47.6 & 43.9 & 58.3 & 63.2 & \textbf{64.8} \\
\bottomrule
\end{tabular*}
\end{table}

On RoboCasa-GR1, LAPT-Scale reaches 64.8\% average success, showing that the
gain also transfers to household tabletop manipulation.

\begin{table}[H]
\centering
\caption{Main downstream policy results on the three original real-robot tasks,
with 20 trials per task. Diffusion Policy uses the same 250-hour robot corpus.}
\label{tab:main_real_robot}
\footnotesize
\setlength{\tabcolsep}{2pt}
\begin{tabular*}{\linewidth}{@{\extracolsep{\fill}}lcccccccccccc@{}}
\toprule
\textbf{Method}
& \multicolumn{4}{c}{$\pi_{0.5}$~\cite{intelligence2025pi_}}
& \multicolumn{4}{c}{JoyAI-RA~\cite{zhang2026joyaira01foundationmodel}}
& \multicolumn{4}{c}{\textbf{LAPT-Scale}} \\
& Mouse & Head. & Cup & Avg. & Mouse & Head. & Cup & Avg. & Mouse & Head. & Cup & Avg. \\
\midrule
\textbf{Success Rate (\%)} & 55.0 & 40.0 & \textbf{75.0} & 56.7 & 80.0 & \textbf{95.0} & 65.0 & 80.0 & \textbf{85.0} & \textbf{95.0} & \textbf{75.0} & \textbf{85.0} \\
\midrule
\multicolumn{13}{l}{Diffusion Policy~\cite{chi2023diffusionpolicy}: Mouse 50.0, Head. 40.0, Cup 45.0, Avg. 45.0.} \\
\bottomrule
\end{tabular*}
\end{table}

The real-robot results test whether human-video latent supervision remains
useful after embodiment alignment. LAPT-Scale achieves 85.0\% average success
over the three tasks, improving over $\pi_{0.5}$ and JoyAI-RA. At the task
level, it improves over JoyAI-RA on Mouse Pad and Cup while matching it on
Headphones. Overall, Tables~\ref{tab:main_robotwin}--\ref{tab:main_real_robot}
suggest that Stage-1 latent-action pre-training can benefit both simulation and
physical robot deployment. We view these results as evidence that LAM-based
human-video pre-training remains compatible with strong downstream policy
training; the scaling behavior is analyzed separately below.
\FloatBarrier

\subsection{Scaling with Annotated Human Video}
\label{sec:scaling}

We next study the central scaling question: whether increasing only the amount
of LAM-annotated human video improves downstream robot performance. The LAM is
always trained on the full 23.3K-hour corpus; we vary only the amount of
cleaned, filtered, and annotated human video used for Stage-1 VLA pre-training,
while keeping robot data, policy architecture, and evaluation protocol fixed.

\begin{figure}[htbp]
    \centering
    \includegraphics[width=\linewidth]{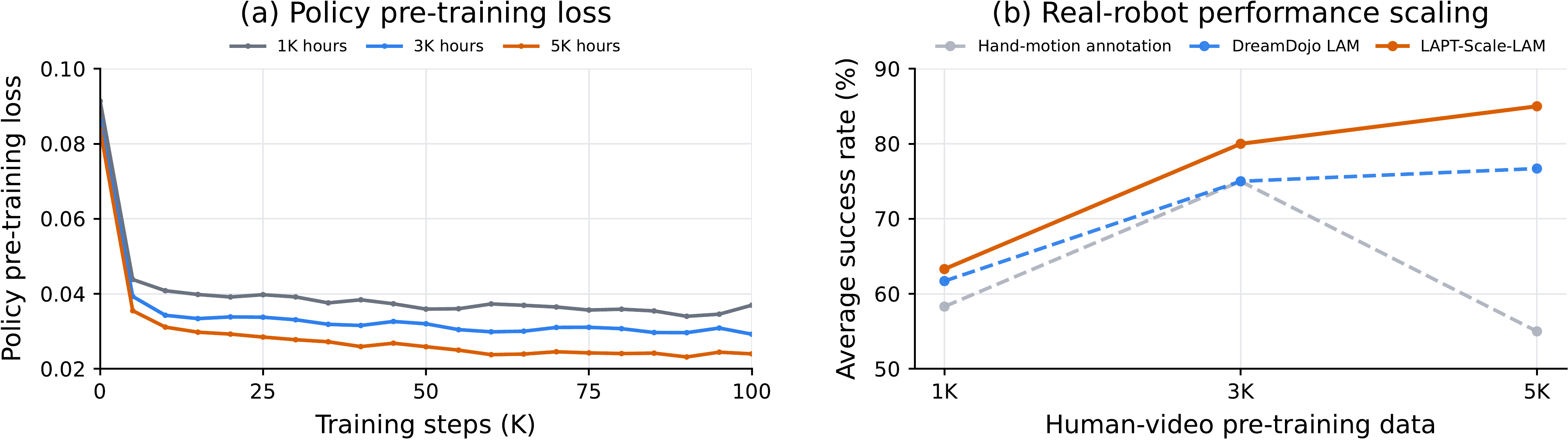}
    \caption{\textbf{Scaling behavior of LAPT-Scale.} Left: policy
    pre-training loss for different amounts of human-video pre-training data.
    Right: real-robot average success rate as the amount of human-video
    pre-training data increases, with all robot data held fixed.}
    \label{fig:lapt_scaling}
\end{figure}

We compare three forms of human-video supervision. Hand-motion annotation uses
explicit hand-derived pseudo-actions, the DreamDojo LAM~\cite{gao2026dreamdojo}
uses its official pretrained reconstruction-based LAM, and LAPT-Scale-LAM uses the full perceptual, semantic, and
motion-aware alignment signals. This comparison tests whether scale helps from more
video alone or from higher-quality action-like supervision. The pretrained
DreamDojo LAM differs from the MSE-only diagnostic trained on our data with
its reconstruction-plus-KL recipe.

Fig.~\ref{fig:lapt_scaling} shows that LAPT-Scale-LAM benefits most
consistently from additional human video, improving from 63.3\% at 1K hours to
80.0\% at 3K hours and 85.0\% at 5K hours. Larger LAM-annotated datasets also
reach lower Stage-1 pre-training losses.

The baselines show why annotation quality matters. Explicit hand-motion
pipelines can accumulate source-dependent hand-estimation noise under camera
motion, occlusion, hand-visibility changes, and viewpoint shifts. A hand tracker
can produce plausible but wrong pseudo-actions in these cases, and downstream
policy training does not correct the labels. Such noise may explain why hand
motion improves from 1K to 3K hours but degrades to 55.0\% at 5K hours.

DreamDojo LAM scales more reliably than hand motion and reaches 76.7\% at 5K
hours, but remains below LAPT-Scale-LAM. This suggests that pixel-level
future-frame reconstruction alone may not fully capture the semantic and motion
structure needed for moving hand-object interactions, motivating the controlled
alignment ablation below. The full per-task breakdown and shared 0-hour control
(55.0\%) are in Appendix~\ref{sec:appendix_scaling_breakdown}.

\subsection{Controlled Stage-1 and Alignment Ablations}
\label{sec:controlled_ablations}

We remove Stage 1 with initialization, robot data, Stage-2/3 schedules, and
evaluation fixed, and extend Stage-2 training in a compute-matched control
(Table~\ref{tab:stage1_controls_main}; Appendix~\ref{sec:appendix_stage1_control}).
Both remain below the full model.

\begin{table}[H]
\centering
\caption{Stage-1 controls: average success on the three original real-robot tasks.}
\label{tab:stage1_controls_main}
\footnotesize
\begin{tabular*}{\linewidth}{@{\extracolsep{\fill}}llc@{}}
\toprule
Stage-1 supervision & Stage-2 training & Success (\%) \\
\midrule
None & Standard & 55.0 \\
None & Extended, compute-matched & 53.3 \\
LAPT-Scale, 5K hours & Standard & 85.0 \\
\bottomrule
\end{tabular*}
\end{table}

We retrain the annotation-to-policy pipeline after removing each LAM alignment
term (Table~\ref{tab:loss_ablation_main}).

\begin{table}[H]
\centering
\caption{Controlled LAM alignment ablation. RT-H is RoboTwin Hard success.}
\label{tab:loss_ablation_main}
\footnotesize
\setlength{\tabcolsep}{4pt}
\begin{tabular*}{\linewidth}{@{\extracolsep{\fill}}lccc@{}}
\toprule
LAM objective & PSNR & LARY & RT-H (\%) \\
\midrule
MSE-only (Recon.+KL) & 33.9844 & 0.675 & 74.2 \\
Full w/o LPIPS & 34.2105 & 0.691 & 82.5 \\
Full w/o DINO & 34.1500 & 0.698 & 84.1 \\
Full w/o flow & 34.3012 & 0.685 & 77.8 \\
Full objective & \textbf{34.4217} & \textbf{0.713} & \textbf{89.3} \\
\bottomrule
\end{tabular*}
\end{table}

Removing flow yields the largest RoboTwin Hard drop (11.5 points), despite
high PSNR, illustrating the limits of pixel fidelity as a diagnostic.

\subsection{Harder Real-Robot Tasks and Generalization}
\label{sec:harder_tasks_generalization}

Four harder tasks cover paper-ball disposal, pen and self-occluded eraser
grasping, and multi-stage book stowing. One joint policy uses 200 demonstrations
per task, also used to post-train $\pi_{0.5}$. We test seen conditions and unseen
position, illumination, and background shifts with 20 trials per task per domain
(protocols and domain images in Appendix~\ref{sec:appendix_robustness}).

\begin{table}[htbp]
\centering
\caption{Four-task generalization success (\%); Average includes all four domains.}
\label{tab:generalization_main}
\footnotesize
\setlength{\tabcolsep}{4pt}
\begin{tabular*}{\linewidth}{@{\extracolsep{\fill}}lccccc@{}}
\toprule
Method & Seen & Position & Illumination & Background & Average \\
\midrule
Diffusion Policy & 60.0 & 20.0 & 15.0 & 15.0 & 27.5 \\
$\pi_{0.5}$ & \textbf{85.0} & 50.0 & 45.0 & 40.0 & 55.0 \\
LAPT-Scale & \textbf{85.0} & \textbf{75.0} & \textbf{70.0} & \textbf{70.0} & \textbf{75.0} \\
\bottomrule
\end{tabular*}
\end{table}

LAPT-Scale and $\pi_{0.5}$ match at 85.0\% in seen conditions, but reach
70.0--75.0\% and 40.0--50.0\%, respectively, under the tested shifts
(Table~\ref{tab:generalization_main}).

\subsection{Human-Video LAM Evaluation}
\label{sec:lary_ablation}

We evaluate the LAM before policy training with two diagnostics: future-frame
reconstruction on human-video transitions and first-person action recognition
on LARY~\cite{nie2026lary} using frozen latent actions with a linear probe.

We compare three LAM variants under the same transition-prediction protocol.
VQ-LAM uses a discrete codebook~\cite{van2017neural}; MSE-only LAM keeps the
continuous latent space but removes the LPIPS, DINO, and optical-flow alignment
terms; LAPT-Scale-LAM uses the full continuous latent space and action-aware
alignment signals. This tests latent-space and alignment choices.

\begin{table}[htbp]
\centering
\caption{Human-video LAM evaluation. Reconstruction metrics are measured on
human-video transitions; LARY reports first-person human-action classification
accuracy from frozen latent actions.}
\label{tab:lam_evaluation}
\begin{tabular}{lcccc}
\toprule
LAM Variant & PSNR $\uparrow$ & SSIM $\uparrow$ & LPIPS $\downarrow$ & LARY Acc. $\uparrow$ \\
\midrule
VQ-LAM & 18.3416 & 0.6741 & 0.3443 & 0.232 \\
MSE-only LAM & 33.9844 & 0.9234 & 0.0769 & 0.675 \\
LAPT-Scale-LAM & \textbf{34.4217} & \textbf{0.9326} & \textbf{0.0546} & \textbf{0.713} \\
\bottomrule
\end{tabular}
\end{table}

Table~\ref{tab:lam_evaluation} shows that LAPT-Scale-LAM improves both
transition reconstruction and latent-action semantics. The gap to VQ-LAM
suggests that discrete codes are too coarse for fine-grained egocentric
hand-object dynamics, while the gain over MSE-only LAM shows the value of
perceptual, semantic, and motion alignment.

\begin{figure}[!htbp]
    \centering
    \includegraphics[width=\linewidth]{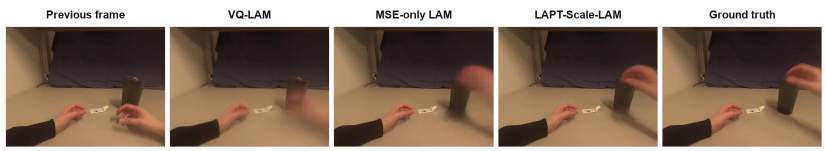}
    \caption{\textbf{Human-video transition reconstruction.} Given the previous
    frame, VQ-LAM, MSE-only LAM, and LAPT-Scale-LAM predict the next frame,
    shown against the ground truth. Additional visualizations are provided in
    the appendix.}
    \label{fig:lam_reconstruction_comparison}
\end{figure}

The qualitative example in Fig.~\ref{fig:lam_reconstruction_comparison}
illustrates the same trend: VQ-LAM produces blurry reconstructions with
noticeable hand-position errors, and MSE-only LAM remains visually blurred.
LAPT-Scale-LAM reconstructs the transition most similarly to the ground truth,
especially around the moving hand and object.

These diagnostics complement downstream evaluation by showing that Stage-1
annotations retain transition information before robot embodiment alignment.

\subsection{Interpreting the Learned Latent Action Space}
\label{sec:interpretability}
\label{sec:qualitative}

To probe what the learned latent action token represents, we perform
counterfactual latent-action replacement on human ego-view videos, which match
the training source of LAPT-Scale-LAM. For each donor-target pair, we keep the
target visual context fixed but replace its latent action with one inferred from
a donor video. We then decode three short-horizon predictions: the donor
self-rollout, the target base rollout, and the target rollout under
donor-latent replacement. This isolates the latent action from scene context:
if the token carries transition information, the replacement should alter the
target rollout without simply copying donor appearance.

\begin{figure}[htbp]
    \centering
    \includegraphics[width=\linewidth]{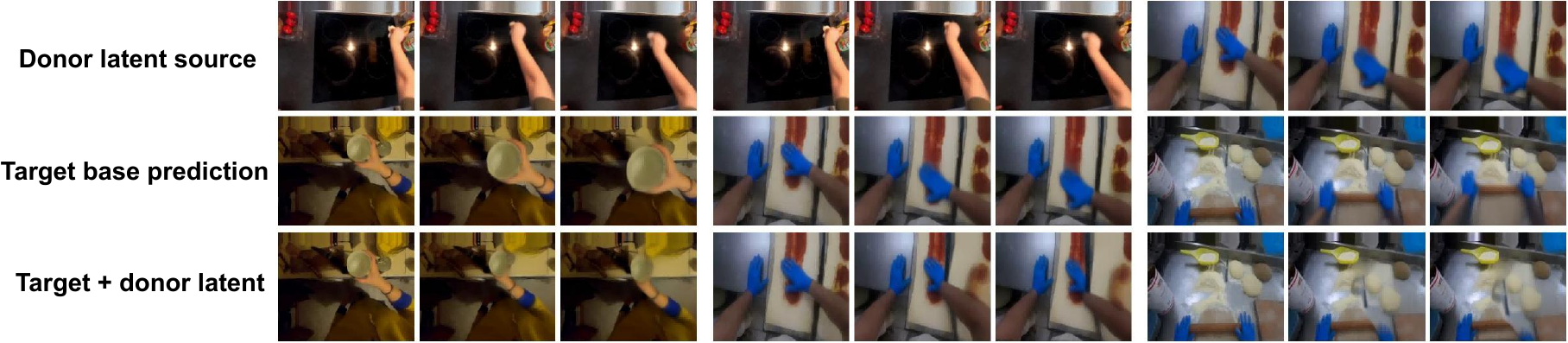}
    \caption{\textbf{Human ego-view latent-action replacement across three donor-target pairs.}
    Rows show donor self-rollout frames, the target base prediction,
    and the target prediction after replacing the target latent action with the
    donor latent action. Across all three examples, the replacement changes the
    short-horizon motion pattern while preserving the overall target scene
    layout.}
    \label{fig:lam_replacement_human_a}
\end{figure}

Across the examples in Figure~\ref{fig:lam_replacement_human_a}, replacement
changes how the target sequence evolves without globally rewriting the target
scene. The clearest effects appear as camera-relative hand-object motion
changes: the donor latent action redirects the short-horizon movement tendency,
while the scene layout remains largely target-specific. These results suggest
that LAM tokens encode transferable information about how the scene changes,
rather than only what the scene contains.


\FloatBarrier
\section{Conclusion and Limitations}
\label{sec:conclusion}

We presented LAPT-Scale, a latent action pre-training paradigm that turns
large-scale egocentric human videos into supervision for embodied policy
learning. By learning continuous transition variables directly from video and
using them for Stage-1 VLA pre-training, LAPT-Scale provides a scalable way to
exploit human videos without explicit robot action labels. Across simulation,
real-robot, scaling, and LAM diagnostic experiments, our results suggest that
LAM-based human-video supervision improves downstream manipulation and becomes
more useful as the amount of annotated human video increases, supporting latent
actions as scalable human-video annotations. Four harder real-robot tasks
also show robustness under the tested position and appearance shifts.

The main limiting assumption is that visual latent actions learned from human
videos can become useful robot-policy supervision after embodiment alignment;
they are not guaranteed to align with independently controllable robot degrees
of freedom. As a result, target-robot embodiment alignment remains necessary.
The approach may also fail when videos contain weak manipulation signal, large
embodiment gaps, heavy occlusion, or visual changes dominated by ego-camera
motion. Our experiments are limited in robot and task coverage: the real-robot
study uses seven tasks and one target platform, with 20 trials per condition.
This limits the precision of success estimates, and broader validation is
needed before claiming deployment robustness. Future work should study stronger
embodiment alignment, longer-horizon latent-action composition, broader robot
coverage, and more systematic scaling laws across human-video data, model size,
and robot demonstrations.

\clearpage
\bibliography{main}  

\clearpage
\appendix
\providecommand{\AppendixImagePrefix}{} 

\section{Human-Video Data Construction}
\label{sec:appendix_data}

\subsection{Source Composition}

LAPT-Scale-LAM is trained on a 23.3K-hour egocentric human-video corpus
collected from six egocentric human-video sources: Egocentric-10K~\cite{buildai2025egocentric10k},
EgoDex~\cite{hoque2025egodex}, Ego4D~\cite{grauman2022ego4d} after
EgoVid-style filtering~\cite{wang2024egovid}, Xperience~\cite{xperience_10m},
EgoVerse~\cite{punamiya2026egoverse}, and self-collected human videos. The
corpus is built by merging complementary sources rather than relying on a
single dataset. Broad daily-activity datasets provide scene and object
diversity, manipulation-focused datasets increase the density of hand-object
interaction, and self-collected videos improve coverage of target-relevant
activities and camera viewpoints.

\begin{table}[H]
\centering
\caption{Human-video source composition. Raw hours denote the source mixture
used to form the 23.3K-hour LAM training corpus. The cleaned subset column
denotes the portion retained after manipulation-focused filtering for Stage-1
policy pre-training.}
\label{tab:appendix_data_sources}
\footnotesize
\setlength{\tabcolsep}{5pt}
\renewcommand{\arraystretch}{1.12}
\resizebox{\textwidth}{!}{%
\begin{tabular}{>{\raggedright\arraybackslash}p{0.18\textwidth}>{\raggedright\arraybackslash}p{0.25\textwidth}>{\raggedright\arraybackslash}p{0.33\textwidth}cc}
\toprule
Source & Primary Content & Role in Corpus & Raw Hours & Cleaned Stage-1 Hours \\
\midrule
Egocentric-10K~\cite{buildai2025egocentric10k}
& Filtered egocentric activity clips
& Additional LAM coverage; excluded from Stage-1 scaling because it lacks
hand-pose annotations needed by the hand-motion baseline
& 10.0K & 0.0K \\
EgoDex~\cite{hoque2025egodex}
& Dexterous egocentric manipulation
& Fine-grained hand-object motion
& 0.7K & 0.5K \\
Ego4D filtered by EgoVid~\cite{grauman2022ego4d,wang2024egovid}
& Ego4D clips retained by EgoVid-style filtering
& Broad daily activity with stronger manipulation relevance
& 1.8K & 0.8K \\
Xperience~\cite{xperience_10m}
& Large-scale egocentric multimodal activity data
& Scene, object, and motion diversity
& 1.1K & 0.4K \\
EgoVerse~\cite{punamiya2026egoverse}
& Diverse egocentric robot-learning videos
& Cross-scene and cross-task diversity
& 1.2K & 0.5K \\
Self-collected human videos
& Egocentric human manipulation videos collected by us
& Target-relevant activities and camera viewpoints
& 8.5K & 2.9K \\
\bottomrule
\end{tabular}
}
\end{table}
\FloatBarrier

\subsection{Filtering, Mixing, and Annotation}

We first standardize all videos to a common decoding format, normalize frame
rates and resolutions for training, remove clips with severe corruption or
missing frames, and segment long videos into short temporal clips. We also
remove near-duplicate clips after segmentation to avoid over-representing
repeated scenes or repeated camera trajectories.

For policy pre-training, we apply an additional manipulation-focused filtering
stage. A clip is retained when it contains visible hand-object interaction and
temporal visual change that can plausibly correspond to an action. We discard
segments dominated by strong blur, extreme ego-camera motion, missing hands,
long static intervals, severe occlusion, or weak manipulation content. This
produces a 5.1K-hour cleaned subset used for the main Stage-1 policy
pre-training run.

The source mixture is constructed in two stages. First, all candidate videos are
standardized and pooled into the 23.3K-hour corpus used to train
LAPT-Scale-LAM. This stage favors coverage: even clips with moderate
manipulation density can help the LAM learn generic egocentric visual dynamics,
camera-relative motion, and object appearance variation. Second, we apply the
stricter manipulation filter described above to build the 5.1K-hour Stage-1
policy pre-training subset. This stage favors supervision quality: clips are
kept when they contain visible hands, object interaction, and meaningful
temporal change.

The 1K-, 3K-, and 5K-hour scaling subsets are sampled from the cleaned pool and
are nested when possible: the 3K-hour subset contains the 1K-hour subset plus
additional clips, and the 5K-hour subset contains the 3K-hour subset plus
additional clips whenever the source manifests allow it. We also cap or
rebalance very large sources during sampling so that scaling primarily
increases the amount of action-like supervision rather than simply increasing
the dominance of one source. In practice, this preserves a mixture of household
manipulation, dexterous-hand interaction, and general egocentric daily activity
across all scales.

For fair comparison with the hand-motion annotation baseline, the cleaned
Stage-1 pool only includes sources where the same filtering and hand-pose-based
pseudo-action pipeline can be applied. Egocentric-10K remains useful for
broadening LAM training coverage, but it does not provide the hand-pose
annotations required by the hand-motion baseline. We therefore exclude it from
the cleaned Stage-1 scaling subsets, which keeps the comparison focused on the
form of human-video supervision rather than source-specific annotation
availability.

We also audit the semantic diversity of the available human-video metadata. The
fine-grained taxonomy identifies 12 scene categories, 18 task categories, 77
scene-task cells, 21 action labels, 54 object labels, and 272 action-object
pairs. We use these statistics as a metadata diversity audit rather than as
exact training-hour composition, because weakly labeled or unlabeled sources
remain part of the large-scale video pool.

After filtering, the frozen LAPT-Scale-LAM encoder annotates each retained clip
with transition-level latent actions. For a clip $x_{1:T}$, the encoder is
applied to adjacent frames to produce posterior means
$\bar{\bm{z}}_{1:T-1}$, which are used as deterministic Stage-1 policy targets.
The latent annotations are stored together with the corresponding visual
context, language or clip-level text when available, source metadata, and masks
for valid temporal positions. Invalid transitions caused by decoding failure or
clip boundaries are excluded from the Stage-1 loss. This annotation format lets
the same policy training code consume human-video latent-action chunks in Stage
1 and executable robot action chunks in Stages 2--3.

\section{Training and Implementation Details}
\label{sec:appendix_training_details}

\subsection{LAM and Policy Training}

The controlled LAM diagnostic variants are trained on the same human-video
corpus with the same clip sampling and optimization protocol.
LAPT-Scale-LAM encodes each video transition
into a continuous latent action and reconstructs the next frame from the current
visual context and latent action. The full model uses the reconstruction,
LPIPS, DINO feature, optical-flow, and KL terms described in the main paper.
The MSE-only diagnostic uses DreamDojo's reconstruction-plus-KL training
recipe on our data, omitting the LPIPS, DINO, and optical-flow alignment terms.
In contrast, the DreamDojo baseline used for Stage-1 scaling uses its official
pretrained LAM. It is not the same checkpoint as the controlled MSE-only
diagnostic. The
VQ-LAM baseline replaces the continuous latent action with a discrete codebook
while keeping the transition prediction protocol fixed.

The LAM uses a patch size of 16, model dimension 1024, 24 Transformer encoder
blocks, 24 Transformer decoder blocks, and 16 attention heads. The latent
action is parameterized as a 32-dimensional continuous Gaussian posterior. The
training objective uses MSE reconstruction as the base term, KL weight
$1\times10^{-6}$, LPIPS weight 0.01, DINO feature weight 0.003, and optical-flow
weight $3.3\times10^{-5}$. DINO alignment uses a frozen DINOv2 ViT-B/14
backbone, and flow alignment uses RAFT-large from the torchvision optical-flow
models.

The LAM training configuration also uses paired augmentation. For each paired
sample, the configuration uses four repeats and resamples the latent action
under augmentation. The color augmentation ranges are 0.3 for brightness,
contrast, and saturation, 0.08 for hue, gamma in $[0.7,1.4]$, Gaussian noise
standard deviation 0.035, blur probability 0.2, and minimum crop scale 0.9.

LAPT-Scale-Policy uses the same VLM backbone and action expert across all
experiments. Stage 1 trains the policy on LAM-annotated human-video clips, where
the target is the continuous latent action produced by the frozen LAM encoder.
Stage 2 continues training the same action expert on 250 hours of target-robot
embodiment-alignment data, replacing the Stage-1 latent target with executable
robot action chunks without introducing a separate action head or adapter. The
action expert supports up to 32 target dimensions; lower-dimensional targets
are padded and excluded from the loss with a mask. Stage 3 further adapts the
same policy on task-specific robot data. Across scaling experiments, the
embodiment-alignment data, downstream task data, policy architecture, and
training schedule are fixed; only the amount and source of human-video
pre-training supervision are changed.

\subsection{Compute and Hyperparameters}

We separate the computational cost into three parts: LAM training, offline
human-video annotation, and policy training. LAM training is a one-time cost on
the 23.3K-hour human-video corpus. After training, the frozen LAM encoder is
used offline to annotate the cleaned 5.1K-hour subset and the nested scaling
subsets. Policy training then consumes the cached latent-action annotations in
Stage 1, so downstream scaling experiments do not require rerunning the LAM
encoder except when changing the annotated human-video pool. Table~\ref{tab:appendix_hyperparams}
summarizes the implementation, training, and compute configuration used in our
experiments.

\begin{table}[!htbp]
\centering
\caption{Implementation, training, and compute details for LAPT-Scale-LAM
training, offline annotation, and policy training.}
\label{tab:appendix_hyperparams}
\footnotesize
\setlength{\tabcolsep}{5pt}
\begin{tabular}{lll}
\toprule
Component & Item & Value \\
\midrule
Human-video corpus & LAM training data & 23.3K hours \\
Human-video corpus & Cleaned Stage-1 subset & 5.1K hours \\
Human-video corpus & Scaling subsets & 1K, 3K, and 5K hours \\
LAM & Latent action dimension & 32 \\
LAM & Latent type & Continuous Gaussian posterior \\
LAM & Prediction target & Next-frame reconstruction \\
LAM & Visual losses & MSE, LPIPS, DINO, optical-flow, KL \\
LAM & Patch size / model dimension & 16 / 1024 \\
LAM & Encoder / decoder depth & 24 / 24 blocks \\
LAM & Attention heads & 16 \\
LAM & Image resolution & $320\times240$ \\
LAM & Input clip length / frame stride & 2 frames / random 1--4-frame gap \\
LAM & Optimizer / learning rate & AdamW / $2.5\times10^{-5}$ \\
LAM & Weight decay / gradient clipping & $10^{-2}$ / 0.3 \\
LAM & Batch size / training steps & 16 per GPU / 1,000,000 \\
LAM & Max epochs & 1000 \\
LAM & Samples per epoch & 10,000,000 \\
LAM & Precision / distributed strategy & bf16 / DDP \\
LAM & Checkpoint / log interval & 5000 / 1000 steps \\
LAM & Padding / sampling strategy & Repeat / manual \\
LAM losses & KL / LPIPS / DINO / flow weights & $10^{-6}$ / 0.01 / 0.003 / $3.3\times10^{-5}$ \\
LAM augmentation & Paired augmentation & Enabled, 4 repeats, resample latent \\
LAM compute & Hardware & 32 H200 GPUs \\
LAM compute & Training time & 154.3 hours at 1.8 steps/s \\
LAM compute & Total GPU hours & 4,938 H200 GPU-hours \\
LAM annotation & Annotated subset size & 5.1K hours \\
LAM annotation & Annotation throughput & 3 video hours / H200-hour \\
LAM annotation & Total annotation cost & 1,700 H200 GPU-hours \\
Policy Stage 1 & Supervision & LAM posterior-mean latent-action chunks \\
Policy Stage 1 & Human-video scale & 1K / 3K / 5K / 5.1K hours \\
Policy Stage 2 & Embodiment-alignment data & 250 hours \\
Policy Stage 2 & Supervision & Target-robot action chunks \\
Policy Stage 3 & Supervision & Task-specific robot action chunks \\
Policy & Action expert capacity & Up to 32 target dimensions \\
Policy & Action chunk horizon & 30 \\
Policy & Optimizer / scheduler & AdamW / cosine with minimum learning rate \\
Policy & Learning-rate groups & base / Qwen-VL interface / action model \\
Policy & Gradient accumulation & 1 \\
Policy compute & Hardware & 32 H200 GPUs \\
Policy compute & Detailed training schedule & Tables~\ref{tab:appendix_policy_pretrain_config}
and~\ref{tab:appendix_real_robot_posttrain_config} \\
\bottomrule
\end{tabular}
\end{table}
\FloatBarrier

\begin{table*}[!t]
\centering
\caption{Policy pre-training and embodiment-alignment configuration. Learning
rates are reported for the base, Qwen-VL interface, and action-model parameter
groups.}
\label{tab:appendix_policy_pretrain_config}
\footnotesize
\setlength{\tabcolsep}{4pt}
\renewcommand{\arraystretch}{1.12}
\begin{tabular}{llccccc}
\toprule
Stage & Frozen modules & Batch size & Base LR & Qwen-VL LR & Action LR & Time \\
\midrule
Stage 1 & None & 1024 (32/GPU $\times$ 32) & $2.5{\times}10^{-5}$ & $1{\times}10^{-5}$ & $1{\times}10^{-4}$ & 43h \\
Stage 2 & Qwen-VL LM backbone & 1024 (32/GPU $\times$ 32) & $2.5{\times}10^{-5}$ & $1{\times}10^{-5}$ & $1{\times}10^{-4}$ & 50h \\
\bottomrule
\end{tabular}
\vspace{2pt}
\begin{minipage}{0.98\textwidth}
\footnotesize Both stages are reported at 100K training steps.
\end{minipage}
\end{table*}

\begin{table*}[!t]
\centering
\caption{Real-robot post-training configuration. All three tasks use full
fine-tuning, task-specific robot demonstrations, and 30K training steps.}
\label{tab:appendix_real_robot_posttrain_config}
\footnotesize
\setlength{\tabcolsep}{6pt}
\renewcommand{\arraystretch}{1.12}
\begin{tabular}{lccccc}
\toprule
Task & Batch size & Base LR & Qwen-VL LR & Action LR & Time \\
\midrule
Mouse Pad & 64 (8/GPU $\times$ 8) & $3{\times}10^{-5}$ & $1{\times}10^{-5}$ & $1{\times}10^{-4}$ & 7h \\
Headphones & 64 (8/GPU $\times$ 8) & $3{\times}10^{-5}$ & $1{\times}10^{-5}$ & $1{\times}10^{-4}$ & 7h \\
Cup & 64 (8/GPU $\times$ 8) & $3{\times}10^{-5}$ & $1{\times}10^{-5}$ & $1{\times}10^{-4}$ & 8h \\
\bottomrule
\end{tabular}
\end{table*}
\FloatBarrier

\section{Baseline Implementation Details}
\label{sec:appendix_baseline_details}

Table~\ref{tab:appendix_baseline_implementation} summarizes the main
implementation differences among the human-video supervision baselines. For
fair comparison, all variants use the same cleaned human-video subsets, the
same Stage-1 policy architecture, the same 250-hour target-robot
embodiment-alignment data, and the same downstream post-training protocol. The
comparison therefore isolates the form and quality of the supervision extracted
from human videos rather than changes in robot data or policy capacity.

\begin{table*}[!t]
\centering
\caption{Implementation comparison of human-video supervision baselines. All
policy baselines share the same VLA backbone, action expert, robot alignment
data, and downstream training schedule.}
\label{tab:appendix_baseline_implementation}
\footnotesize
\setlength{\tabcolsep}{3pt}
\renewcommand{\arraystretch}{1.12}
\begin{tabular}{>{\raggedright\arraybackslash}p{0.18\textwidth}>{\raggedright\arraybackslash}p{0.24\textwidth}>{\raggedright\arraybackslash}p{0.35\textwidth}>{\raggedright\arraybackslash}p{0.16\textwidth}}
\toprule
Method & Target / Latent Type & Training Signal & Used For \\
\midrule
Hand-motion annotation
& Reconstructed hand trajectory; explicit pseudo-action with hand pose
& Hand-pose trajectory fitting
& Stage-1 policy scaling \\
DreamDojo LAM
& Implicit transition action; latent action without hand pose
& Pixel-level future-frame reconstruction
& Stage-1 policy scaling \\
VQ-LAM
& Implicit transition action; discrete codebook without hand pose
& Future-frame reconstruction with vector quantization
& LAM diagnostic baseline \\
MSE-only LAM
& Implicit transition action; continuous Gaussian without hand pose
& MSE future-frame reconstruction and KL
& LAM diagnostic baseline \\
LAPT-Scale-LAM
& Implicit transition action; continuous Gaussian without hand pose
& MSE, LPIPS, DINO, optical-flow, and KL alignment
& Stage-1 policy scaling and LAM diagnostics \\
\bottomrule
\end{tabular}
\end{table*}
\FloatBarrier

For the explicit annotation baseline in
Table~\ref{tab:appendix_scaling_real_breakdown}, we convert egocentric human
videos into action-like supervision through reconstructed hand motion.
Specifically, this baseline treats the human hand as a dexterous end-effector
and uses hand-pose trajectories as pseudo-action targets for human-video
pre-training. It uses the same clip sampling, filtering, policy architecture,
robot data, and downstream training schedule as LAPT-Scale-LAM. This comparison
is intended to isolate the effect of replacing explicit hand-motion supervision
with LAM-derived implicit actions.

The hand-motion baseline depends on explicit hand reconstruction, which can
produce noisy pseudo-actions under occlusion, motion blur, unusual viewpoints,
or contact-rich manipulation. The LAM baselines avoid explicit hand-pose
estimation by learning transition variables directly from video. DreamDojo LAM
and MSE-only LAM primarily test reconstruction-driven latent actions, VQ-LAM
tests whether a discrete bottleneck is sufficient for egocentric manipulation,
and LAPT-Scale-LAM tests whether continuous latents with perceptual, semantic,
and motion-aware alignment produce better action-like supervision.

\section{Evaluation Protocol Details}
\label{sec:appendix_eval_protocol}

For all downstream robot evaluations, we report task success rate. A rollout is
counted as successful only when the task-specific goal condition is satisfied
within the evaluation horizon; otherwise it is counted as a failure. For
simulation benchmarks, we follow the benchmark-provided success detectors and
episode termination rules. For real-robot tasks, success is judged by the final
task state after execution and uses the same criteria for all compared methods.

\subsection{Robot-Data Quantities and Evaluation Accounting}
The 250-hour Stage-2 corpus is heterogeneous embodiment-alignment data,
comprising 162,767 episodes and 26.7M frames rather than per-task data. For
Stage 3, we use 50 demonstrations per original real-robot task, 200 per added
hard task, 50 clean plus 500 randomized demonstrations per RoboTwin task, and
1,000 demonstrations per RoboCasa task. The original tasks use 20 evaluation
trials per task. The four harder tasks use 20 trials per task per domain, or
320 trials per method across the seen domain and three shifted domains.

\subsection{Simulation Benchmarks}

RoboTwin 2.0 evaluation follows the benchmark Easy and Hard settings. For
RoboTwin post-training, each task uses 50 clean demonstrations and 500 heavily
randomized demonstrations. The Easy setting evaluates policies under the
standard benchmark initialization distribution, while the Hard setting
introduces stronger variation in object poses and scene layouts. We report the
average success rate across the benchmark task suite for each setting; the main
LAPT-Scale policy reaches 90.0\% in Easy and 89.3\% in Hard. All compared
methods are evaluated with the same task split, observation interface, action
interface, and success definition.

RoboCasa-GR1 evaluation uses the tabletop manipulation setting with 24 tasks and
50 rollouts per task. We follow the benchmark-provided simulator reset
distribution and task success checks, and report the average success rate across
all RoboCasa rollouts. The reported RoboCasa-GR1 number averages over the
evaluated task instances rather than selecting the best task-specific
checkpoint; the main LAPT-Scale policy reaches 64.8\% average success.

\subsection{Real-Robot Tasks}

Real-robot evaluation is conducted on an AgiBot G1 robot after target-robot
embodiment alignment. We evaluate three tasks: Mouse Pad, Headphones, and Cup.
Each task is evaluated over 20 trials per method. Between trials, the scene is
reset to the task-specific initial distribution, and all methods are evaluated
under the same reset and success criteria. A Mouse Pad trial succeeds if the
robot completes the requested manipulation of the mouse pad; a Headphones trial
succeeds if the headphones are manipulated to the target state; and a Cup trial
succeeds if the cup task reaches the specified final configuration.

\begin{table}[!htbp]
\centering
\caption{Summary of downstream evaluation protocols.}
\label{tab:appendix_eval_protocol}
\footnotesize
\setlength{\tabcolsep}{4pt}
\begin{tabular}{lcccc}
\toprule
Benchmark & Setting & Metric & Evaluation scope & Success Criterion \\
\midrule
RoboTwin 2.0 & Easy & Success rate & Official task suite & Benchmark detector \\
RoboTwin 2.0 & Hard & Success rate & Official task suite & Benchmark detector \\
RoboCasa-GR1 & Tabletop & Success rate & 24 tasks $\times$ 50 rollouts & Benchmark detector \\
Real robot & Mouse Pad & Success rate & 20 & Task completion \\
Real robot & Headphones & Success rate & 20 & Task completion \\
Real robot & Cup & Success rate & 20 & Task completion \\
\bottomrule
\end{tabular}
\end{table}
\FloatBarrier

\pagebreak[3]
\section{Additional Real-Robot Scaling Results}
\label{sec:appendix_scaling_breakdown}

\begin{table}[htbp]
\centering
\caption{Stage-1 scaling on the original real-robot tasks. Each entry is an
average success rate (\%) over three tasks with 20 trials per task. The shared
0-hour column is the same robot-only control.}
\label{tab:scaling_with_zero}
\footnotesize
\begin{tabular*}{\linewidth}{@{\extracolsep{\fill}}lcccc@{}}
\toprule
Stage-1 supervision & 0 hours & 1K hours & 3K hours & 5K hours \\
\midrule
Hand motion & 55.0 & 58.3 & 75.0 & 55.0 \\
DreamDojo LAM & 55.0 & 61.7 & 75.0 & 76.7 \\
LAPT-Scale-LAM & 55.0 & 63.3 & 80.0 & 85.0 \\
\bottomrule
\end{tabular*}
\end{table}

\FloatBarrier

\begin{table}[!htbp]
\centering
\caption{Per-task breakdown for the real-robot scaling study. Human-video scale
denotes the amount of annotated human video used for VLA pre-training; all
robot data is held fixed. Each task is evaluated over 20 trials.}
\label{tab:appendix_scaling_real_breakdown}
\begin{tabular}{lccccc}
\toprule
Human Supervision & Human Video & Mouse Pad & Headphones & Cup Task & Avg. \\
\midrule
Hand-motion annotation & 1K hours & 45.0 & 80.0 & 50.0 & 58.3 \\
Hand-motion annotation & 3K hours & 80.0 & 45.0 & 100.0 & 75.0 \\
Hand-motion annotation & 5K hours & 75.0 & 50.0 & 40.0 & 55.0 \\
DreamDojo LAM & 1K hours & 60.0 & 55.0 & 70.0 & 61.7 \\
DreamDojo LAM & 3K hours & 80.0 & 85.0 & 60.0 & 75.0 \\
DreamDojo LAM & 5K hours & 65.0 & 75.0 & 90.0 & 76.7 \\
LAPT-Scale-LAM & 1K hours & 50.0 & 70.0 & 70.0 & 63.3 \\
LAPT-Scale-LAM & 3K hours & 70.0 & 75.0 & 95.0 & 80.0 \\
LAPT-Scale-LAM & 5K hours & \textbf{85.0} & \textbf{95.0} & 75.0 & \textbf{85.0} \\
\bottomrule
\end{tabular}
\end{table}
\FloatBarrier

\subsection{Stage-1 Removal Control}
\label{sec:appendix_stage1_control}

The robot-only control removes Stage 1 while retaining the same initialization,
250-hour embodiment-alignment corpus, Stage-2/3 data, and evaluation protocol.
The compute-matched control extends Stage-2 training to match the training
compute budget. Unlike the Diffusion Policy comparison, Stage-1 removal tests
human-video supervision within the same policy architecture.

\begin{table}[!htbp]
\centering
\caption{Stage-1 removal and compute-matched controls on real-robot tasks.}
\label{tab:appendix_stage1_control}
\footnotesize
\begin{tabular}{lcc}
\toprule
Stage-1 supervision & Stage-2 training & Average success (\%) \\
\midrule
None (robot-only) & Standard & 55.0 \\
None (robot-only) & Extended, compute-matched & 53.3 \\
LAPT-Scale, 5K hours & Standard & 85.0 \\
\bottomrule
\end{tabular}
\end{table}
\FloatBarrier

\section{Additional LAM Diagnostics}
\label{sec:appendix_lam_diagnostics}

\subsection{Reconstruction Protocols}
\label{sec:appendix_lam_reconstruction_protocols}

In addition to the single-step reconstruction results reported in the main
paper, we evaluate each LAM variant under two rollout protocols. The first is a
\textit{teacher-forced next-frame reconstruction} protocol: for every time step,
the model receives the ground-truth current frame $x_t$ and predicts
$\hat{x}_{t+1}$. This measures whether the latent action extracted from a local
transition preserves enough information to reconstruct the next visual state
without accumulating temporal errors. The second is an
\textit{autoregressive rollout} protocol: the model receives the ground-truth
first frame $x_1$, predicts $\hat{x}_2$, and then conditions on its own
prediction to generate subsequent frames. This evaluates temporal stability and
whether reconstruction errors compound over multiple latent-action steps.

For this quantitative comparison, we include LAPA~\cite{ye2024latent} and
VILLA-X~\cite{chen2025villa} as additional latent-action model baselines.
LAPA represents a discrete latent-action annotation approach, while VILLA-X
represents a video latent-action baseline
trained for egocentric transition prediction. Together with MSE-only LAM, these
baselines test whether the gains come from the LAPT-Scale alignment design
rather than only from using a latent transition model.

\begin{figure}[H]
    \centering
    \includegraphics[width=0.92\linewidth] {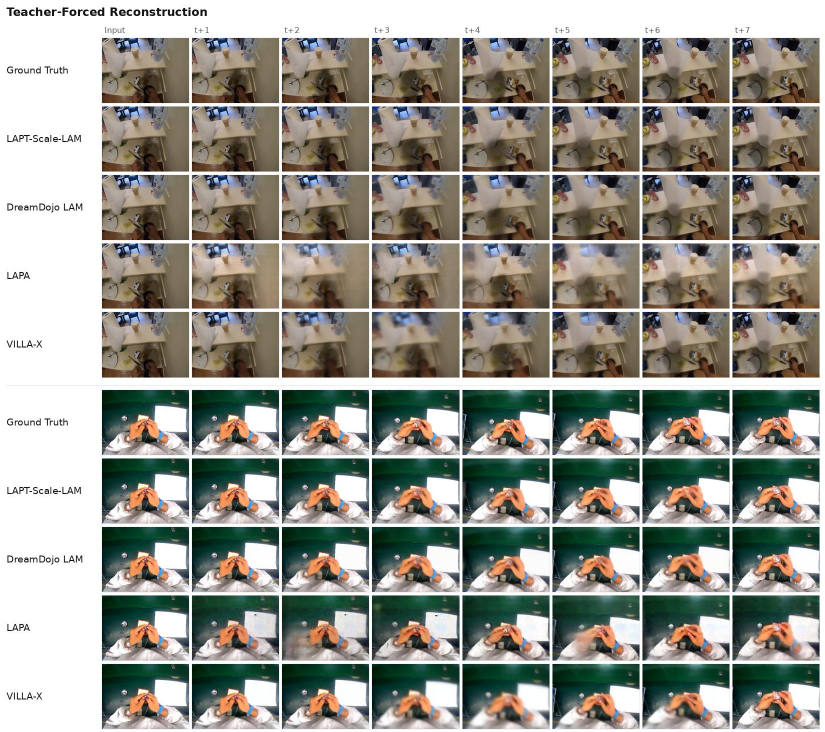}
    \caption{\textbf{Teacher-forced LAM reconstruction comparison.} Each
    prediction conditions on the ground-truth previous frame. LAPA and VILLA-X
    show weaker local transition fidelity, MSE-only LAM is smoother around hand
    contacts, and LAPT-Scale-LAM preserves sharper hand-object structure.}
    \label{fig:appendix_lam_teacher_forced}
\end{figure}

\begin{figure}[!htbp]
    \centering
    \includegraphics[width=0.92\linewidth] {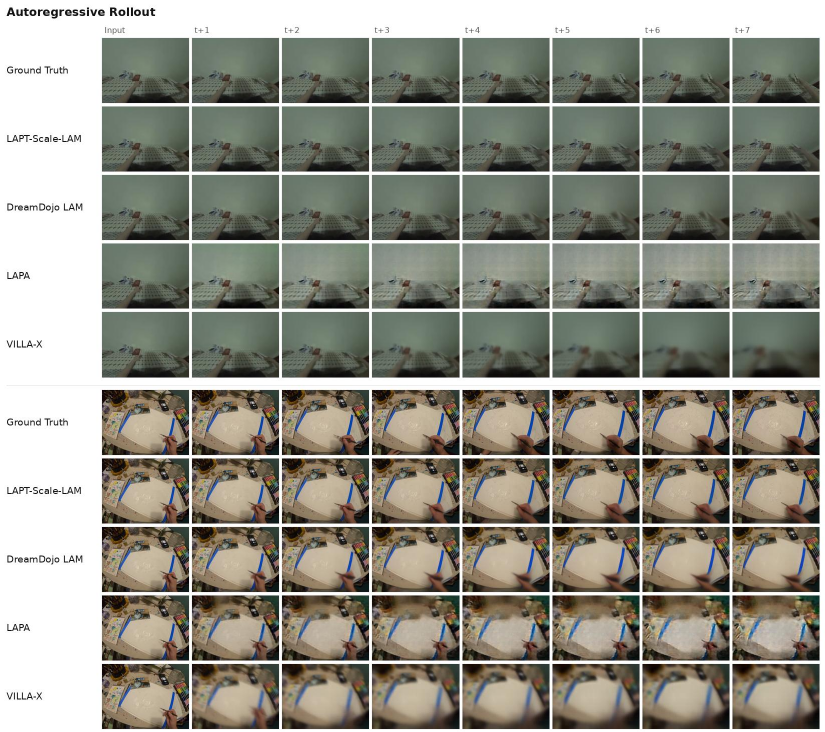}
    \caption{\textbf{Autoregressive LAM rollout comparison.} Later predictions
    condition on the model's own generated frames. Error accumulation makes the
    differences more visible: MSE-only LAM gradually loses fine hand structure,
    while LAPT-Scale-LAM preserves a visible hand region and more coherent
    short-horizon object motion.}
    \label{fig:appendix_lam_autoregressive}
\end{figure}

The metrics in Table~\ref{tab:appendix_lam_rollout_protocols} are computed over
3500 frames sampled from a randomly held-out 100-hour evaluation split of our
mixed human-video dataset. Qualitative examples for both protocols are shown in
Figs.~\ref{fig:appendix_lam_teacher_forced} and
\ref{fig:appendix_lam_autoregressive}.

\begin{table}[!htbp]
\centering
\caption{LAM reconstruction comparison under teacher-forced and autoregressive
rollout protocols. Teacher-forced evaluation conditions each prediction on the
ground-truth previous frame, while autoregressive evaluation starts from the
ground-truth first frame and then conditions on the model's previous prediction.
Higher PSNR and SSIM are better; lower LPIPS is better.}
\label{tab:appendix_lam_rollout_protocols}
\footnotesize
\setlength{\tabcolsep}{4pt}
\begin{tabular}{llccc}
\toprule
Protocol & LAM Variant & PSNR $\uparrow$ & SSIM $\uparrow$ & LPIPS $\downarrow$ \\
\midrule
Teacher-forced & LAPA & 18.3416 & 0.6741 & 0.3443 \\
Teacher-forced & VILLA-X & 26.9595 & 0.7973 & 0.1685 \\
Teacher-forced & MSE-only LAM & 33.9844 & 0.9234 & 0.0769 \\
Teacher-forced & LAPT-Scale-LAM & \textbf{34.4217} & \textbf{0.9326} & \textbf{0.0546} \\
\midrule
Autoregressive & LAPA & 15.8607 & 0.5551 & 0.4593 \\
Autoregressive & VILLA-X & 22.5951 & 0.7008 & 0.3228 \\
Autoregressive & MSE-only LAM & 28.3290 & 0.8426 & 0.1768 \\
Autoregressive & LAPT-Scale-LAM & \textbf{28.8291} & \textbf{0.8510} & \textbf{0.1347} \\
\bottomrule
\end{tabular}
\end{table}
\FloatBarrier

These two protocols expose the differences progressively. In the
teacher-forced setting, LAPA and VILLA-X already show weak local transition
fidelity: hand-object boundaries are less stable and the predicted motion is
often visually inconsistent with the ground-truth next frame. MSE-only LAM is
substantially stronger under teacher forcing because every prediction is
conditioned on a clean ground-truth frame, but it still tends to produce
smoother hands and object contacts. LAPT-Scale-LAM gives the sharpest
teacher-forced reconstructions, while DreamDojo-style reconstruction remains
slightly blurrier around the moving hand and contacted objects.

Fig.~\ref{fig:appendix_lam_teacher_forced} illustrates the easier
teacher-forced case. Because every prediction receives a clean previous frame,
the comparison mainly reflects how well each latent action preserves the local
transition. The strongest differences appear around the hand and the object
being manipulated, where reconstruction-only objectives tend to smooth contact
boundaries.

\FloatBarrier

The autoregressive setting makes this gap much more visible. Once each model
must condition on its own previous prediction, small contact and hand-position
errors compound over time. LAPA and VILLA-X drift quickly, while MSE-only LAM
gradually loses fine hand structure; by the final generated frames, the hand is
often no longer clearly recognizable. LAPT-Scale-LAM remains more stable under
the same rollout protocol, preserving a visible hand region and coherent
short-horizon object motion for longer. This supports the view that the
additional perceptual, semantic, and motion alignment terms improve temporal
latent-action structure rather than only single-step reconstruction quality.

Fig.~\ref{fig:appendix_lam_autoregressive} evaluates the same models under a
stricter rollout protocol. Here, the model must reuse its own generated frames,
so small hand-position errors and blurred contacts become part of the next
input. This is why the gap between MSE-only LAM and LAPT-Scale-LAM is more
visible in later frames than in the teacher-forced comparison.

\FloatBarrier

\subsection{Similar Motions Across Different Scenes}
\label{sec:appendix_similar_motion_different_scene}

To test whether the learned latent action is tied to visual appearance or to
transition-level motion, we compare clips that contain similar short-horizon
actions in different scenes. For each query transition, we retrieve nearest
neighbors in the LAPT-Scale-LAM latent-action space while excluding clips from
the same source video. This produces groups where the background, object
identity, lighting, and hand appearance can differ substantially, but the
short-horizon motion pattern is similar.

\begin{figure}[!htbp]
    \centering
    \includegraphics[width=0.92\linewidth] {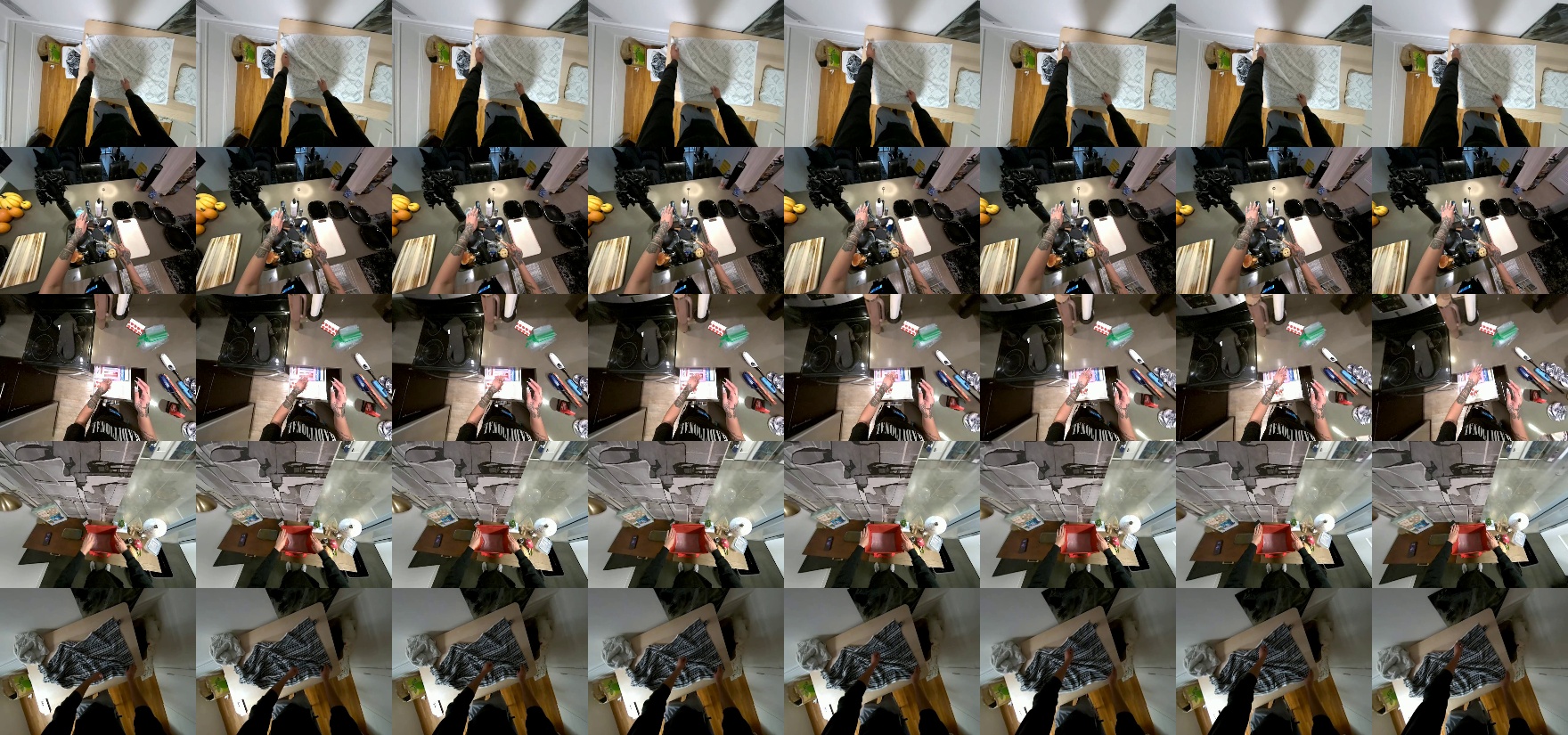}
    \vspace{2pt}

    \includegraphics[width=0.92\linewidth] {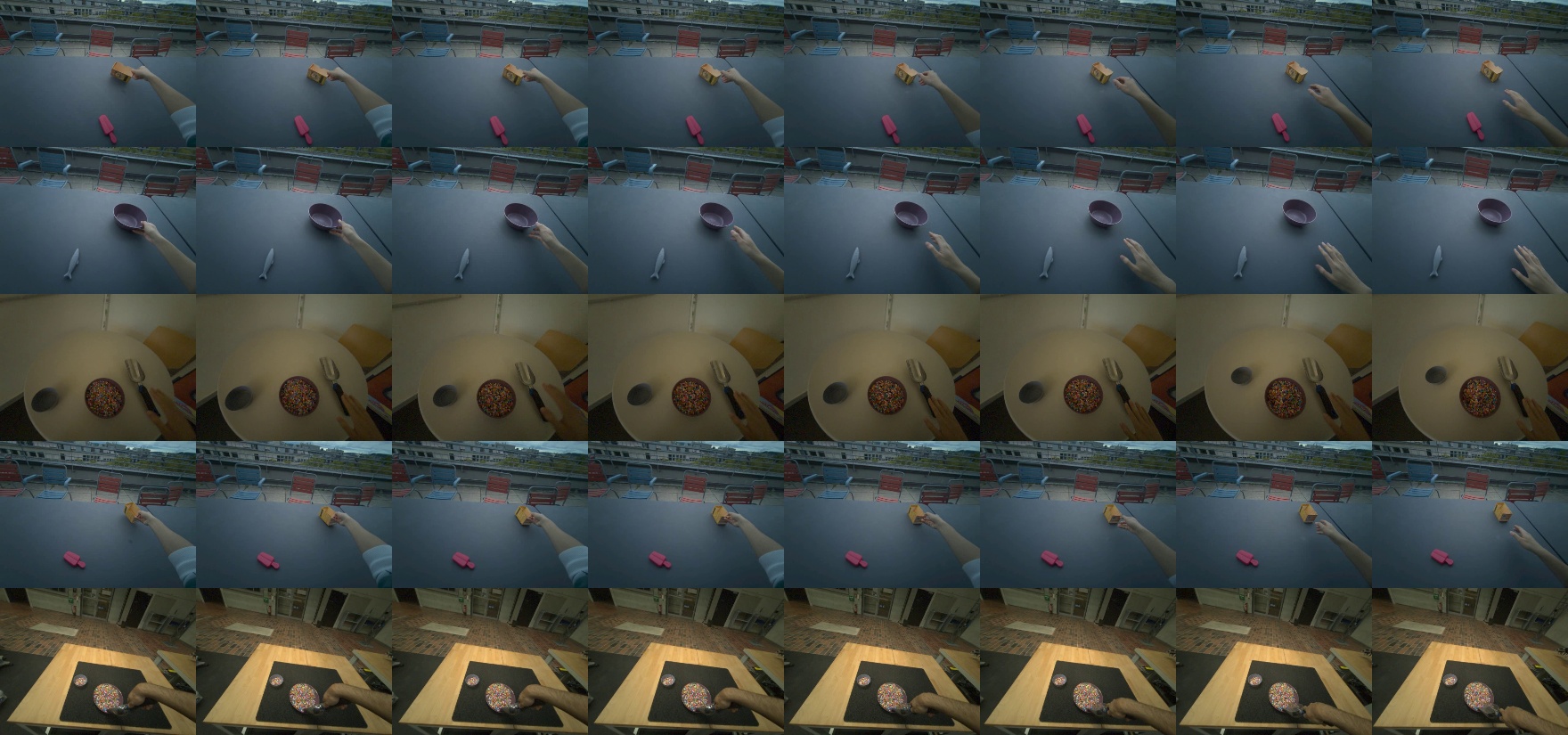}
    \caption{\textbf{Latent-action clusters with shared motion patterns across
    different data sources.} Each panel shows examples grouped by the learned
    latent-action representation. Clusters are computed from 50K
    motion-filtered 16-frame windows after PCA reduction to 64 dimensions and
    50-way clustering. Although the visual scenes and objects vary, samples
    within a cluster exhibit similar short-horizon motion patterns, suggesting
    that LAPT-Scale-LAM captures transition-level structure beyond
    dataset-specific appearance.}
    \label{fig:appendix_motion_clusters}
\end{figure}

For the cluster visualization, we sample 50K motion-filtered windows from
2.36M candidate windows across 11,981 episodes. Each window contains 16 frames
with stride 16. We retain windows above the 70th percentile of the motion score
(threshold 4.805), reduce latent-action features to 64 PCA dimensions
(explained-variance sum 0.7756), and cluster the resulting features into 50
groups. The selected clusters in Fig.~\ref{fig:appendix_motion_clusters}
illustrate repeated hand-object motion patterns across visually different
clips.

This analysis complements the latent-action replacement experiment reported in
the main paper. Nearest-neighbor retrieval asks whether
similar motions naturally cluster together across scenes, while replacement
asks whether a latent action inferred from one scene can change the rollout in
another scene without copying the donor appearance. Together, these diagnostics
support the interpretation that the LAM token encodes how the scene changes
rather than only what the scene looks like.

The selected clusters in Fig.~\ref{fig:appendix_motion_clusters} are not meant
to show near-duplicate frames. Instead, they show that clips with different
backgrounds, objects, and hand appearances can still be close in latent-action
space when their short-horizon motion is similar. This is the desired behavior
for Stage-1 supervision: the policy should learn reusable transition patterns
rather than memorize source-specific visual appearance.

\FloatBarrier

\section{Stage-1 Latent Rollout Diagnostics}
\label{sec:appendix_stage1_rollouts}

To further diagnose Stage-1 human-video pre-training, we evaluate whether the
learned latent-action space is meaningful enough for decoder rollout and
whether the VLA policy can predict action-relevant latent variables from
context. We compare two rollout settings: using the ground-truth latent action
encoded by LAPT-Scale-LAM and using the latent action sampled from the Stage-1
VLA policy. The ground-truth-latent rollout tests whether the compressed latent
space preserves enough transition information for decoding, while the VLA
rollout tests whether Stage-1 pre-training learns to predict latent actions
that still produce coherent human motion.

\begin{table}[!htbp]
\centering
\caption{Stage-1 latent-action rollout quality on human videos. Higher PSNR
and SSIM are better; lower LPIPS is better.}
\label{tab:appendix_stage1_rollout_metrics}
\begin{tabular}{lccc}
\toprule
Rollout Target & PSNR $\uparrow$ & SSIM $\uparrow$ & LPIPS $\downarrow$ \\
\midrule
VLA-predicted latent rollout & 25.5371 & 0.8137 & 0.3033 \\
Ground-truth latent rollout & \textbf{30.7705} & \textbf{0.8982} & \textbf{0.2271} \\
\bottomrule
\end{tabular}
\end{table}
\FloatBarrier

Table~\ref{tab:appendix_stage1_rollout_metrics} shows that ground-truth latent
actions can reconstruct future visual transitions with strong fidelity,
indicating that the compressed LAM latent space retains meaningful motion
information. More importantly, the VLA-predicted latent rollout also achieves
reasonable reconstruction quality, showing that Stage-1 pre-training learns
action-related latent structure rather than only static appearance cues.

The two examples below visualize this diagnostic at different manipulation
scales. The dice example emphasizes short, fast hand-object motion, while the
sandwich example contains longer contact sequences and larger appearance
changes. In both cases, the Stage-1 policy-predicted latents produce rollouts
that preserve the dominant motion pattern even though they are not used as
executable robot actions.

\begin{figure}[!htbp]
    \centering
    \includegraphics[width=0.92\linewidth] {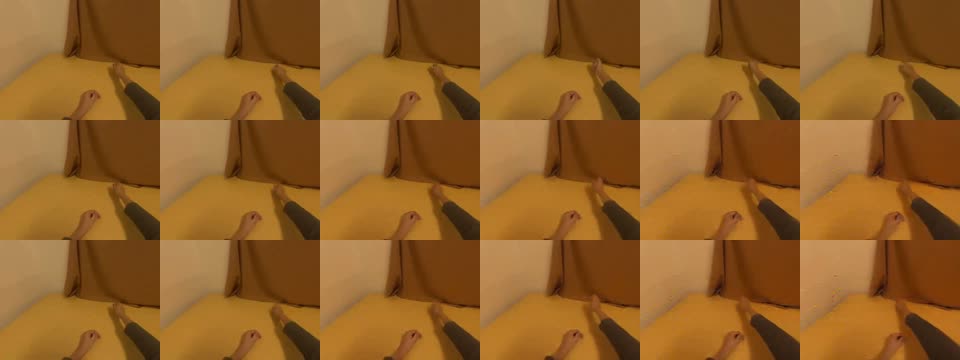}
    \vspace{2pt}

    \small \textit{Gather the four dice with one hand and roll them on the
    brown background while sitting.}
    \caption{\textbf{Stage-1 latent-action rollout on human videos: dice
    rolling.} Frames are
    sampled every six video frames and arranged from left to right within each
    row. The first row shows the ground-truth video, the second row shows
    decoder rollout using the ground-truth latent action encoded by
    LAPT-Scale-LAM, and the third row shows decoder rollout using the latent
    action predicted by the Stage-1 VLA policy.}
    \label{fig:appendix_stage1_rollout_dice}
\end{figure}

The sandwich sequence is visually harder because multiple ingredients and hand
contacts change across the clip. The rollout remains useful as a diagnostic
because it tests whether predicted latent actions encode the transition
tendency rather than merely reconstructing a static scene.

\begin{figure}[!htbp]
    \centering
    \includegraphics[width=0.92\linewidth] {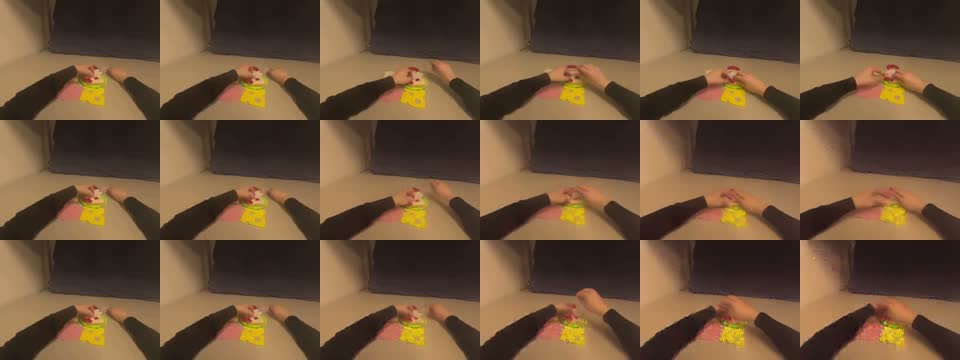}
    \vspace{2pt}

    \small \textit{Make a sandwich using breads, cheese, onion, ketchup, mayo,
    tomato, lettuce, and ham while sitting at a table.}
    \caption{\textbf{Stage-1 latent-action rollout on human videos: sandwich
    making.} Frames are sampled every six video frames and arranged from left
    to right within each row. The Stage-1 predictions produce rollouts with
    motion patterns close to the ground-truth-latent rollout and the observed
    video, indicating that human-video pre-training learns meaningful human
    motion representations rather than only static visual appearance.}
    \label{fig:appendix_stage1_rollout_sandwich}
\end{figure}
\FloatBarrier

\section{Additional Robustness Evaluation}
\label{sec:appendix_robustness}

\subsection{Harder Real-Robot Tasks}
To test whether the gains extend beyond the three original tabletop tasks, we
add paper-ball disposal, slender-pen grasping, small eraser grasping under
self-occlusion, and multi-stage book stowing. The first task requires disposal
into a target receptacle; pen and eraser require precision grasping despite
small contact regions and partial occlusion; book stowing requires a controlled
edge push, grasp, vertical reorientation, and shelving. We jointly train one
policy with 200 demonstrations per task. The same downstream demonstrations
are used to post-train $\pi_{0.5}$ from its official checkpoint.

\subsection{Generalization to Unseen Conditions}
We evaluate the four harder tasks under the training distribution and three
unseen shifts: object position, illumination, and background/tablecloth
appearance. Each entry averages 20 trials per task and domain. LAPT-Scale and
$\pi_{0.5}$ are tied in-distribution, whereas LAPT-Scale degrades less under all
three shifts.

\begin{figure}[!htbp]
  \centering
  \includegraphics[width=0.96\linewidth] {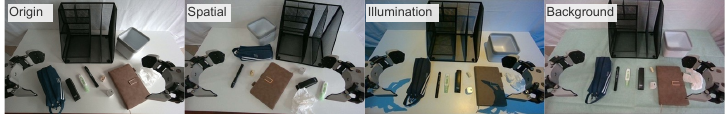}
  \caption{Original evaluation domain and the three unseen generalization shifts.}
  \label{fig:appendix_generalization_shifts}
\end{figure}

\begin{table}[!htbp]
\centering
\caption{Four-task real-robot generalization (20 trials per task per domain).}
\label{tab:appendix_generalization}
\footnotesize
\setlength{\tabcolsep}{4pt}
\begin{tabular}{lccccc}
\toprule
Method & Seen & Position & Illumination & Background & Average \\
\midrule
Diffusion Policy & 60.0 & 20.0 & 15.0 & 15.0 & 27.5 \\
$\pi_{0.5}$ & 85.0 & 50.0 & 45.0 & 40.0 & 55.0 \\
LAPT-Scale & 85.0 & 75.0 & 70.0 & 70.0 & 75.0 \\
\bottomrule
\end{tabular}
\end{table}

\end{document}